# A Hindi Speech Actuated Computer Interface for Web Search

Kamlesh Sharma
Research Scholar, Dept. of CSE
Lingaya's University
Faridabad, Haryana, India

Dr. S.V.A.V. Prasad
Dean of R&D
Lingaya`s University
Faridabad, Haryana, India

Dr. T. V. Prasad
Dean of Computing Sciences
Visvodaya Technical Academy
Kavali, Andhra Pradesh, India

*Abstract*— **Aiming at increasing system simplicity and flexibility, an audio evoked based system was developed by integrating simplified headphone and user-friendly software design. This paper describes a Hindi Speech Actuated Computer Interface for Web search (HSACIWS), which accepts spoken queries in Hindi language and provides the search result on the screen. This system recognizes spoken queries by large vocabulary continuous speech recognition (LVCSR), retrieves relevant document by text retrieval, and provides the search result on the Web by the integration of the Web and the voice systems. The LVCSR in this system showed enough performance levels for speech with acoustic and language models derived from a query corpus with target contents.**

## I. INTRODUCTION

Information contained on the World Wide Web is inaccessible to many people. The web is primarily a visual medium that requires a keyboard and mouse to navigate, and this disenfranchises several types of users. People who lack in skills to use a keyboard and mouse find navigation difficult. Handicapped users also cannot use keyboard and mouse and people who do not have access to an Internet-capable computer have difficulty even accessing the Web. Speech recognition and generation technologies offer a potential solution to these problems by augmenting the capabilities of a web browser.

In the present development of human computer interaction (HCI), Automatic Speech Recognition interface (ASRI) is an emerging technology for offering users a totally new way of mouse, keyboard and joystick control, by using speech [1]. ASRI systems can provide both typing and web browsing for the disabled who cannot use real mouse and keyboard to obtain information from Internet as normal person do. HSACIWS users can input the query in the form of speech on the web page of the browser by using headphone. After confirmation, HSACIWS can capture the uniform resource locator (URL) and open the web page they are interested in. This novel application integrates functions of character input and cursor control, and may help those disabled to obtain information by using the search engines. World Wide Web as a repository of information for unconstrained and wide dissemination, information is now broadly available over the internet and is accessible from remote sites.

## II. NEED OF INTERFACE

As network and internet have become popular and used by everyone around the word for accessing the data stored in storage mediums attached to network and servers on the internet. Data could be text, images, video, audio and other representation and non representation information. Search engines and directories were used in making queries and searches of stored data then returning the result of the query to the user. Currently used search engines and directories Google, Yahoo, Alta Vista, Ask, MSN Search, AOL, Lycos, Looks Smart and other search engine provide their search service via servers connected to networks and internet. [2]

The search engine primary mechanism is to navigate to a web page requested by user when the page is stored on any server in internet. The users manually type characters words or phrases known as a query into the search engine form. No search engine currently offers an implementation solution allowing user to make queries by speaking the query term in Hindi voice, converting the Hindi speech words into data then converting them to English word query and finally processing the query to perform a search.[3]

Speech recognition technology was developed over the past three decades. It is used in many fields like automatic speech recognition directory, military, defense, medical science, bio-informatics, home automation systems, word processing system, dictation system, embedded systems, query processing and many more systems developed for handicapped persons [7]. Most of the systems work on English language and they function at variable level of effectiveness due to limitation of the software understanding and the complexity and the variability of human speech. Large numbers of methods have been developed to acquire the accuracy in English language.

In order to enable a wider proportion of the population of India to benefit from the internet, there is a need to provide additional interface between the user and machine. Speech being a natural mode of communication among human beings can also be a convenient mode of interaction with a computer. Internationally, efforts are already on to combine hypertext navigation with spoken language. This is of particular significance in India where the rate and level of literacy are quite low. Coupled innovatively with visuals, speech and sound can add a new dimension for conveying information to the common man. It is desirable that the human-machine interface permits communication in one's native language. This is an important issue in a multi-lingual country such as India where large numbers of people are comfortable with Hindi language. If the human oriented information over the internet can be access in Hindi language, the computer can process such hypermedia document and provide the information appropriately to a large number of users [14].

In India, Hindi speech recognition is upcoming field of research and people are working in many aspect of Hindi speech recognition. The Hindi speech engine could be utilized in search engine for querying will make the learning and use very easy for the common man. Such type of interfaces can be used by handicapped also and most importantly the person need not be aware about how to use computers.

### III. SYSTEM DESIGN

The manner in which users interact with a program is known as its user interface. The user interface controls how data is entered and how information is displayed. There are mainly five types of user interfaces: character user interface (CUI), graphical user interface (GUI), touch user interface (TUI), voice user interface (VUI) and brain signal user interface (BSUI). Herein a voice user interface comes into existence. Fig. 2 shows a flowchart of Hindi speech actuated computer interface for web search shows the system design of speech actuated interface.

As shown in Fig. 1, the speech actuated computer interface consists of wearable headset and a laptop computer or desktop computer. Headphone is used to take the speech query from user. A good quality headphone acquired, amplified and digitalized the Audio signal, then wirelessly or transmitted it to the computer. The software running on laptop computer analyzed the signal and performed the instructions issued by the user. The voices utterances are of words fed to statistical speech recognition model using Hidden Markov model (HMM) where the word that were utters most likely are determined. A database was constructed with a list of word defining specific subject like fruits, vegetable, news, recipes, stock, weather etc. [9]

The uttered words are compared to the database words, if uttered word match is found a set of keywords are formed to make a query. This query is input to a search engine and the search engine processing the query and returns the result.

A user would use a speech actuated user interface in which user query is received through a device in signal form of speech and converts these speech signals to digital signals. The digital signals received as input to speech recognition module accepts the natural continuous speech patterns and generates most probable words uttered by using the HMM. The output from speech recognition module is searched against a large database of words stored in previously formulated and trained database. The output from speech recognition module is Hindi words send as input to "Speech conversion from Hindi to English" module which converts these Hindi words into corresponding English words.

The strings of English words is passed to a "Find out keywords and drop the stop constant" module for processing as in Fig. 2. Keywords are identified and marked and that recognized speech which was not useful or will not be used for search query were dropped. The keywords passed to "Make a Query" module formed a new query with the keyword and passed to "Search engine module". In case keywords are not found then the query was passed again through the user Hindi query module. The search engine searched the pages on the internet and fetched the pages on the client machine and finally displayed the results on user's screen. In case results were not found then it displayed a message on user's screen suitably.

The forming of user query was basically a combination of words that were identified from utterance using HMM. HMM [5] was used for representing speech units. Forty eight phone-like acoustic-phonetic units were used to represent Hindi sentences. Only monophone models were used due to paucity of sufficient speech data. The plosives were represented as two units-the first representing the closure, and the second representing the rest of the stop consonant. While distinction was maintained between the releases of aspirated and unaspirated plosives, no such distinction was retained between their closures. In addition to the Hindi phonemes, a few commonly occurring English vowels such as /ae/ were included. At the model level, gender identification technique can be used and the test feature vector sequence can be matched with a gender dependent HMM. Initialisation of emission probability densities of HMM states using a segmented and labelled speech database [6] should lead to better models.

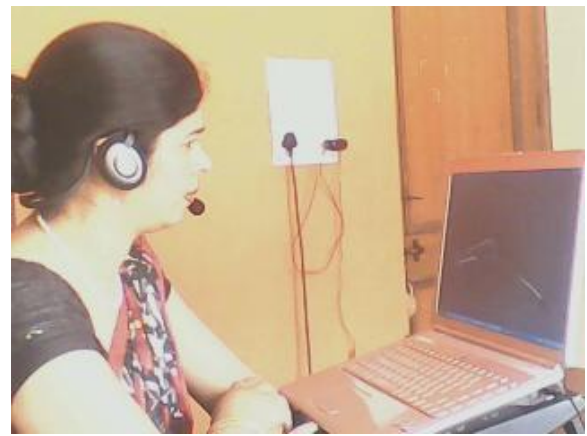

Fig. 1 User interacting with system without mouse and keyboard

Dropping of stop constant and identification of key words was very important to the searching process, as shown in Fig. 2. Search engine returned results on the basis of query formed with the help of keywords identified by find keywords module.

For instance, if user gave a query for processing to the HSACIWS is "*Aaj Dili ki mandi mein aalu ka bhav kya hai*". This query was given as user query in Hindi module. The recognized string of words send as input to speech conversion from Hindi to English module which converts the Hindi string to corresponding string with the help word model and morphological analyzer is "what is the price of potatoes in the market of Delhi today". This converted string is passed as input to find the keywords and dropped the stop constant module which tokenized the keywords like "today", "what", "is", "the", "price", "of", "mango", "in", "Delhi", etc. and drop the stop constants. These keywords were passed as input to make a "Make a query" module which applied preprocessing techniques to generate a suitable query for search engine.

This was for HSACIWS to make sure that the input query was right as recognition was error prone. The system needed to identify the errors and came up with appropriate strategies to overcome the errors. A query was identifying in different ways as shown in Table 1. The given query "*Aaj Dili ki mandi mein Aalu ka Bhav kya hai*" could be recognized as:

Table 1. Query recognition with different variation in keywords

| |
|---|
| Aaj Bili ki mandi mein Aalu ka Bhav kya hai |
| Aaj Dili ki dandi mein Aalu ka Bhav kya hai |
| Aaj Dili ki mandi mein Balu ka Bhav kya hai |
| Aaj Dili ki mandi mein Balu ka Bhav kya hai |
| Aaj Bili ki mandi mein Balu ka Bhav kya hai |
| Aaj Dili ki mandi mein Aalu ka kya hai |
| Aaj Dili ki mandi mein ka Bhav kya hai |

In such type of cases when keywords were not recognized correctly, control was passed back to user query in Hindi module and the user repeated the query again.

The results were displayed but in many cases was not sufficient to answer the user. At this point the user was allowed to optionally select to run a different module to mark hyper text links or filter information from hyper links related to the user query. Depending on the user's choice, the results could then be displayed as text on the screen or played as speech. The output can be seen on screen in two ways, viz., (a) hyperlink number technique as shown in Fig. 3 or (b) employing a data filtering algorithm to get the actual results [10].

In hyperlink number technique a visual numbering was provided to the hyperlinks and indicating to the user how to activate them. The user can activate the hyperlink number module by providing visual number to the module [7]. Data is filtered and processed using a data filtering algorithm according to the user queries and results are sent to the speech synthesizer which translates the result from text to speech before playing it on the client system. [4]

The following are the major issues involved in the development of HSACIWS:

- *Noisy environment:* The target users of HSACIWS were primarily the farmers in rural and semi urban areas. The quality of speech signal was affected by the distance of microphone, environment where system was placed, speech codec's and communication channel. HSACIWS system was expected to work in noisy environments, including background speech.

- *Dialect/Pronunciation variation:* The user spoke in different styles. Each dialect differed from the other at phonetic, phonological, morphological, grammatical and lexical levels. Some time moods (anger, illness, happiness and sadness) of the users effected the dialect/pronunciation variation.

- *Unstructured conversation:* The target audience of the HSACIWS may not have interacted with a computer based information access system. Hence, the conversation was typically unstructured and was filled with inconsistencies including repeats and false starts.

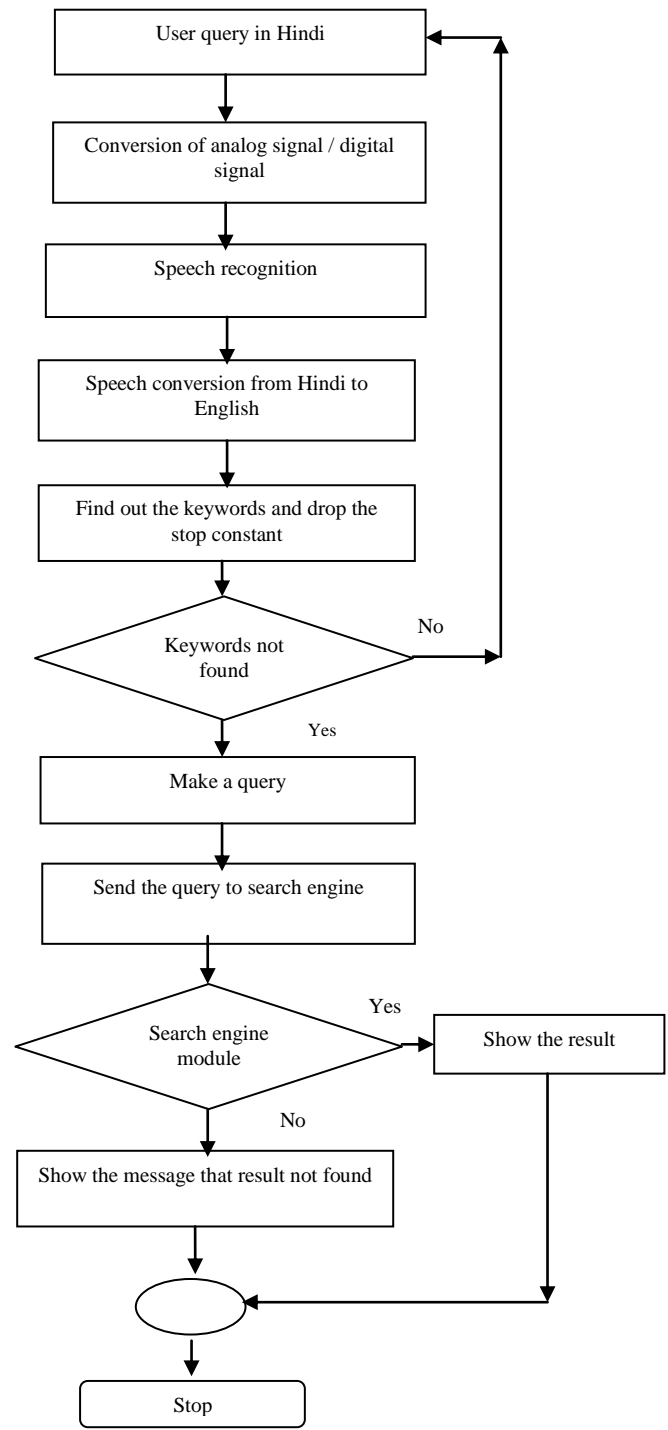

Fig. 2 Flowchart of Hindi speech actuated computer interface for web search

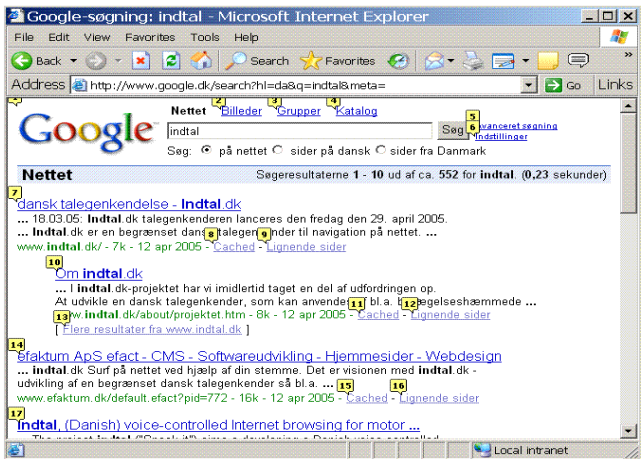

Fig. 3 Displaying hyper links with number in the web Browser [8]

## A. Training Module

The system was trained by actual speakers articulating word continuously. Continuous speech was marked by sounds or phonemes that were connected to each other. The audio signal was processed and features extracted. The signal was smoothed by using different filters to form feature vector which were computed periodically, say every 10 to 20 milliseconds. Many types of features were used including time and frequency masking, tasking of inverse Fourier transforms resulting in a mathematical series of which the coefficient were retained as feature vector. The features were handled mathematically as vectors to simplify the training and recognition computation.

Fig. 4 describes an approach to trained the system for reading the spoken words used in the training were listed in a lexicon and a phonetic word model by using the HMM were from lexicon and phonetic spelling. These HMM word models were iteratively compared to the training speech. Training speech was produced by these HMM words models and the grammar was established with the lexicon a single probabilistic grammar for the sequence of phonemes was formed and stored in dictionary.

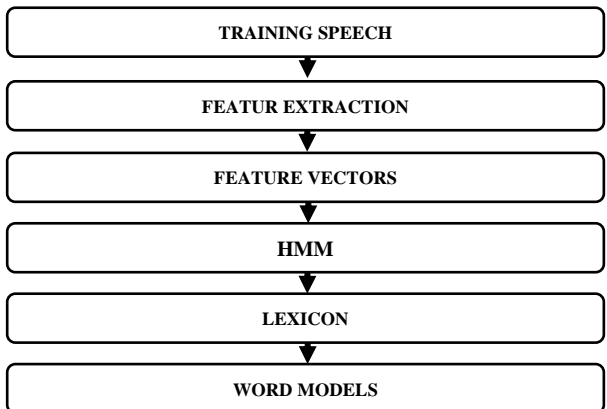

Fig. 4. Functional block diagram of the Training Module

## B. Speech Recognition Module

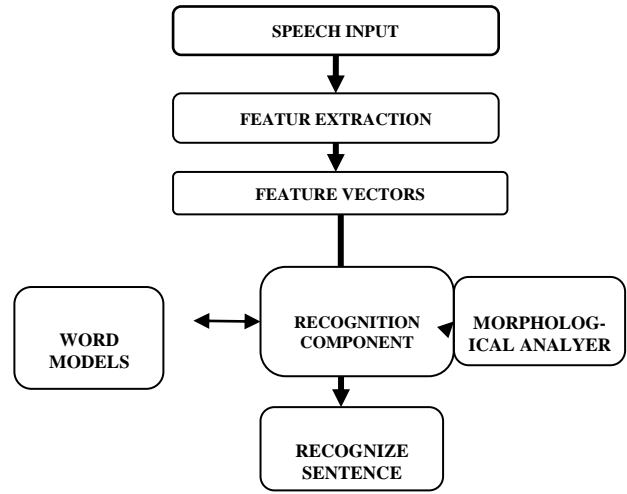

Fig. 5. Functional block diagram of the Speech Recognition Module

Recognition of an unknown speech begins with extracting feature and generating a feature vector for a particular speech by using different vectors generating technique as discussed in training module. Output of feature vector was passed to recognition component as shown in Fig. 5. Recognition component used HMM model sequences allowed by the grammar were searched to find the word sequence in the word model with the highest probability of generating that particular sequence of feature vector.

## C. Morphological Analyzer

Morphological analyer component was used to generate a complete sentence with the help of words recognized by recognized component using the reverse and forward methodology. For translation from one language to other, the source language was first analyzed for finding the required attributes. In the source sentence the words may exist in any of their forms, so we first found their root words and then other attributes. Finding root words in the source language is called Reverse Morphology, also known as Morphological Analyzer. For the target language, the words from the given root word and their attributes were generated, and hence called forward morphology. The Morphological Analyzer constitutes of following sub modules: Input Module, Input Normalizer Module, and Tagger Module, as shown in Fig. 6 and which are described below: [13]

a) *Input Normalizer:* Input Normalizer separates the entered text into the words. It separates the words and stores them whenever blank space is encountered and provides it as input to other modules. This module also searches for the presence of the auxiliary words in the sentence and removes them, if present.

The auxiliary and its attributes are stored for further processing. The presence of auxiliary is language dependent as some languages like Sanskrit does not have auxiliaries. This module only removes the extra spaces between the words and sends the normalized sentence for further processing. [11]

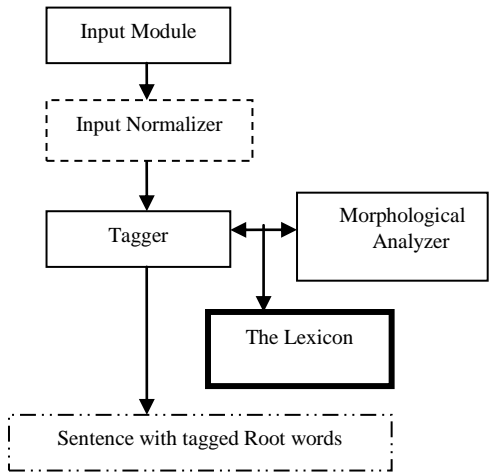

Fig. 6. Functional block diagram of the Morphological Analyzer

b) *Tagger:* The tagger takes the normalized sentence from the previous module and subdivided into it lexical items. Lexical items are not necessarily single words. More than one word in the input sentence may form a lexical item (e.g. give up, put off, etc). The process of dividing the sentence into lexical items is often known as Lexical Analysis. Given "What is the price of potato in Delhi market today" as input, this module would tokenize the words into an array of lexical items. Once the lexical items are obtained the next task of the tagger is to obtain the category and subcategory information for each of these items. It uses morphological analyzer and source lexicon to obtain the category and subcategory information for the lexical items.

c) *Source Language Lexicon:* The bilingual lexicon took Hindi as the source language and English as the target language. The lexicon contained the categorization and sub categorization information about source and target words, as in Table 2. For example:

Table 2. Categorization information

| Category | Subcategory information stored in lexicon |
|---|---|
| Noun | gender, number, person, noun case |
| Verb | tense, aspect, gender |

A given word may carry different categories and subcategories information both in the source as well as in the target language. Further, there may not be exact matching of attributes from the source language to the target language. For instance, the verb in Hindi language has two numbers (singular, plural) and two genders (masculine, feminine) whereas English language verbs does not have gender. The lexicon of a language is its vocabulary, including its words and expressions. More formally, it is a language's inventory of lexemes.

The lexicon includes the lexemes used to actualize words. Lexemes are formed according to morpho-syntactic rules and express sememes. In this sense, a lexicon organizes the mental vocabulary in a speaker's mind: First, it organizes the vocabulary of a language according to certain principles (for instance, all verbs of motion may be linked in a lexical network) and second, it contains a generative device producing (new) simple and complex words according to certain lexical rules. There are five tables a) Hindi root lexicon, b) Hindi verb feature, c) morphological lexicon, d) noun feature and e) suffices. Usually a lexicon is a container for words belonging to the same language. [12]

d) *Morphological Analyzer:* This module is a part of tagger which finds out category and sub-categorical information of the lexical items. The Morphological Analyzer is an integral part of any Natural Language Processing system, especially in the context of Indian languages. For fixed word order languages, the semantics of a word are primarily governed by its absolute and relative position inside a sentence. However, for free word-order languages, any clues about the semantics cannot be obtained from its position in the sentence. In case of Indian languages, which are mostly free order (like Hindi); the semantics (part of speech and other subtleties) are heavily dependent on the surface structure of the word. The task of the morphological analyzer is to identify the structural components of a word, and hence glean information about it. [13]

When an input sentence was fed for translation, the morphological analyzer identified the proper words in the sentence and retrieved necessary information about those words from the lexical database. Lexical database stored only the root form of words and its syntactic and semantic information. With the help of paradigm files, root word was extracted from the original word and all the information about that word was retried.

IV. RESULTS AND DISCUSSIONS

A. *Training and Testing Data Scenario for Experiments*

In order to compare the effectiveness of the HSACIWS system under a scenario of truly limited data resources. There is a collection of training and testing data for the HSACIWS system that contained extremely limited amounts of data. The Training and testing data was extracted from the Indian urban and semi urban areas that have been collected through the questionnaires, Templates and personal talk. The large number of data has been collected through the questionnaires where the users were asked for different question for which the system could train and test.

The limited data consisted of the following resources:

a. *Data Corpus:* The database contains 478 word-aligned phrases and sentences from the user in urban and semi

urban areas. This elicited data collection includes both the training and testing phrases and sentences.

b. *Small Hindi-to-English Lexicon:* The database contained 2390 Hindi to English translation pairs in database. The "Speech conversion from Hindi to English" module of the system used the database and runs each Hindi input word through the morphological analyzer. The Morpher returns the root form of the word, along with a set of morphological features. The root form is then matched against the lexicon, and the set of lexical transfer rules that match the Hindi root are extracted. The morphological features of the input word are then unified with the feature constraints that appear in each of the candidate lexical transfer rules, pruning out the rules that have inconsistent features and generate the English sentence.

c. *Small English-to-Hindi Lexicon:* The database contained 2105 English to Hindi pairs in database. The "Show result" module uses the database to generate the result from English to Hindi.

### B. Experimental Testing

Experiments were conducted to evaluate the baseline system and the improved HSACIWS. The HMM Toolkit was used in this experiments. A set of 100 sentences were randomly chosen from the set of 478 training sentences for test the HSACIWS. A set of 20 subjects were randomly chosen from 100 subjects (students, farmers, housewives and teachers) and were asked to raise a query to the HSACIWS to retrieve the result accordingly and show the user. The user were not trained or provided any information. Initially the experiment is done on pre define queries. The performance of the HSACIWS was calculated by using the percentage performance formula Accuracy which is defined as

$$ \text{Accuracy} = \frac{S - E_s - E_d}{S} \times 100\% $$

where S, $E_s$ and $E_d$ denotes the total number of sentences in the test sentences, the number of substitution errors, and deletion error respectively. The Accuracy as 100 times the ratio of the number of complete sentences recognized correctly to the total number of sentences in the test suite and show the result.

The success of each trial was based on whether the system was able to retrieve the required information to the user or not. For example user will raises the queries to the HSACIWS and system respond accordingly.

User: Sone ka Bhav kya hain.

System: Sone ka Bhav 31500 rupai hai.

User: WHO ka kya matlab hai.

System: WHO ka matlab World Health Organization.

User: Bharat ki Rajdhani kya hai.

System: Bharat ki Rajdhani delhi hai.

Over 200 experiments were conducted on the HSACIWS with 20 users (10 male and 10 female) on 100 different queries. The results are as shown in Table 3 and Table 4:

Table 3. Performance analysis of HSACIWS

| Data Type | Number of Sentences | User | Sex | Total | Accuracy |
|---|---|---|---|---|---|
| Training | 478 | 12 | M/F | 5736 | 79.3% |
| Test | 100 | 10 | M | 1000 | 78.9% |
| Test | 100 | 10 | F | 1000 | 77.8% |

Table 4. Overall system performance analysis

| System | Accuracy |
|---|---|
| HSACIWS | 78.5% |

The success of system was based on the confidence whether the system was able to retrieve the required information correctly to the user or not.

## V. CONCLUSION

We present initial efforts for utilizing spoken Hindi language as a means of communicating web information to common Indians. We presented a voice query retrieval system in Hindi applied to document search on Internet or network.

The results of this experiment suggest that native Indians who are not able to use the computer and/or lack English skills will be able to use voice based control to navigate and obtained responses from the World Wide Web. They will not need to train the speech recognition software to their specific voice.

This system provides access to digital content over the internet to illiterate, vision-impaired, urban and semi-urban Indian people who are not able to read/write English language.

The same Speech interface can be enhanced to work for different regional languages like Punjabi, Marathi, and Telugu, etc. to enhance or to extend it for different regional languages what all is needed is the transfer or translation rules of grammar, which can be generated with great ease by using the same dataset with different target languages. Hence, there is a need to design and develop special user interfaces for accessing web information by speech of different languages.

## AUTHORS' PROFILE

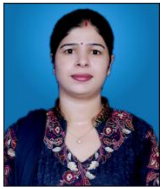

**Ms. Kamlesh Sharmsa** received her masters in Computer Sc. & Engg. degree from Maharshi Dayanand University, Rohtak, India in 2009. She is currently associated with at Lingaya's University, Faridabad in the Dept. of Comp. Sc. & Engg. as Research Scholar. She has over 7 years of teaching experience at under graduate and graduate levels. Her areas of interest are artificial intelligence, operating systems, web mining, Database Management Systems, etc.

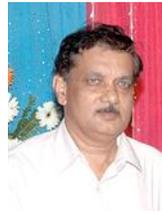

**Dr. S. V. A. V. Prasad** has over 30 years of experience in industry and academics. He has received his master's degree in Electronics & Communications Engg. from Andhra University, AP, India. He earned PhD from Andhra University, Waltair, Visakhapatnam, India. He was with leading research and manufacturing companies in New Delhi, India. He also taught at leading institutions like the Delhi College of Engg. (now Delhi Technological University), Delhi for many years.. He has worked as Head of the Department of Electronics & Communications Engg., Dean of Academic Affairs and as Dean of R&D and Industrial Consultancy at Lingaya's University, Faridabad. He has lectured at various forums on subjects related to electronics, communications, audio engineering, signal processing, etc. Prof. Prasad is a member of IEEE, ISTE, etc. His research interests include audio engineering, signal processing, etc.. He has large number of papers in different journals and conferences.

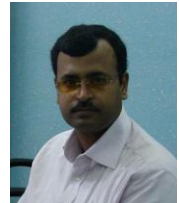

**Dr. T. V. Prasad** has over 17 years of experience in industry and academics. He received his graduate and master's degree in Computer Science from Nagarjuna University, AP, India. He was with the Bureau of Indian Standards, New Delhi for 11 years as Scientist/Deputy Director. He earned PhD from Jamia Millia Islamia University, New Delhi in the area of computer sciences/bioinformatics. He has worked as Head of the Department of Computer Science & Engineering, Dean of R&D and Industrial Consultancy and then as Dean of Academic Affairs at Lingaya's University, Faridabad. He is with Visvodaya Technical Academy, Kavali as Dean of Computing Sciences. He has lectured at various international and national forums on subjects related to computing. Prof. Prasad is a member of IEEE, IAENG, Computer Society of India (CSI), Indian Society of Remote Sensing (ISRS) and APBioNet. His research interests include bioinformatics, artificial intelligence (natural language processing, swarm intelligence, robotics, BCI, knowledge representation and retrieval). He has over 75 papers in different journals and conferences, and also has six books and two chapters to his credit.